\documentclass[conference]{IEEEtran}
\IEEEoverridecommandlockouts
\usepackage{cite}
\usepackage{amsmath,amssymb,amsfonts}
\usepackage{graphicx}
\usepackage{textcomp}
\usepackage{xcolor}
\usepackage{multirow}
\usepackage{url}
\usepackage{algorithm}
\usepackage{algpseudocode}

\usepackage{subcaption}

\usepackage[hidelinks]{hyperref}

\usepackage{tikz}
\usepackage{pgfplots}
\pgfplotsset{compat=1.18} % or 1.17 if your TeX is older

\usetikzlibrary{patterns,arrows.meta}

\def\BibTeX{{\rm B\kern-.05em{\sc i\kern-.025em b}\kern-.08em
    T\kern-.1667em\lower.7ex\hbox{E}\kern-.125emX}}
\begin{document}

\title{Fast NF4 Dequantization Kernels for Large Language Model Inference
}

\author{\IEEEauthorblockN{Xiangbo Qi}
\IEEEauthorblockA{
\textit{University of Southern California}\\
Los Angeles, USA \\
qim@usc.edu}
\and
\IEEEauthorblockN{Chaoyi Jiang}
\IEEEauthorblockA{
\textit{University of Southern California}\\
Los Angeles, USA \\
chaoyij@usc.edu}
\and
\IEEEauthorblockN{Murali Annavaram}
\IEEEauthorblockA{
\textit{University of Southern California}\\
Los Angeles, USA \\
annavara@usc.edu}
}
%\author{}

\maketitle

\begin{abstract}
Large language models (LLMs) have grown beyond the memory capacity of single GPU devices, necessitating quantization techniques for practical deployment. While NF4 (4-bit NormalFloat) quantization enables 4$\times$ memory reduction, inference on current NVIDIA GPUs (e.g., Ampere A100) requires expensive dequantization back to FP16 format, creating a critical performance bottleneck. This paper presents a lightweight shared memory optimization that addresses this gap through principled memory hierarchy exploitation while maintaining full ecosystem compatibility. We compare our technique against the open-source BitsAndBytes implementation, achieving 2.0--2.2$\times$ kernel speedup across three models (Gemma 27B, Qwen3 32B, and Llama3.3 70B) and up to 1.54$\times$ end-to-end improvement by leveraging the 12--15$\times$ latency advantage of shared memory over global memory access. Our optimization reduces instruction counts through simplified indexing logic while using only 64 bytes of shared memory per thread block, demonstrating that lightweight optimizations can deliver substantial performance gains with minimal engineering effort. This work provides a plug-and-play solution for the HuggingFace ecosystem that democratizes access to advanced models on existing GPU infrastructure.
\end{abstract}


\section{Introduction}

Large language models (LLMs) such as GPT-5~\cite{gpt5_release}, Llama 4~\cite{llama4_release}, Claude 4~\cite{claude4_report}, and Qwen3~\cite{qwen3_release} have demonstrated remarkable capabilities across diverse domains, driving widespread adoption in production applications. These applications span code generation~\cite{codex_code_generation, alphacode_competitive_programming}, question answering~\cite{webgpt_browsing, instructgpt_alignment}, and conversational AI systems~\cite{lamda_dialogue}. The exponential growth from GPT-3's 175B parameters~\cite{gpt3_language_generation} to current models exceeding 400B parameters~\cite{llama4_release} has created a deployment challenge where models exceed single GPU memory capacity, necessitating sophisticated quantization techniques~\cite{dettmers2023qlora}.

NF4 (4-bit NormalFloat) quantization~\cite{dettmers2023qlora} has emerged alongside other methods such as GPTQ~\cite{gptq_quantization} and AWQ~\cite{awq_quantization} as the dominant approach for memory-efficient fine-tuning and inference, reducing model footprint by 4$\times$ while maintaining accuracy within 0.1\% of full precision on downstream tasks. However, the current NVIDIA Ampere architecture lacks native 4-bit compute support, requiring dequantization from NF4 to FP16 format for every matrix multiplication during inference. This creates a fundamental performance bottleneck that becomes increasingly severe as model sizes grow and throughput demands increase. Each layer in modern LLMs requires millions of dequantization operations, with the overhead accumulating across hundreds of transformer layers. Our profiling in Table \ref{tab:baseline_overhead} shows that the Qwen3-32B dequantization process accounts for roughly 30--40\% of the end-to-end latency, clearly identifying dequantization as the dominant bottleneck.

Our work addresses this gap through a principled optimization of the dequantization bottleneck using lightweight shared memory techniques. Unlike existing approaches that require complex kernel fusion or offline preprocessing~\cite{park2024lutgemm, guo2024flute}, our method achieves considerable performance gains through focused optimization of memory access patterns while maintaining full compatibility with the HuggingFace Transformers~\cite{huggingface_transformers} and BitsAndBytes ecosystem~\cite{dettmers2023qlora}. The simplicity of our approach enables immediate deployment in production systems and research environments without disrupting existing workflows or requiring model conversion. Experimental results show significant improvements in inference latency and throughput. In summary, our contributions are as follows:
\begin{itemize}
    \item We identify and characterize the dequantization bottleneck in current NF4 implementations through systematic profiling, revealing that memory access patterns rather than computational complexity dominate performance.
    \item We design and implement a lightweight shared memory optimization that leverages the 12--15$\times$ latency advantage of on-chip memory over global memory on Ampere architecture.%~\cite{abdelkhalik2022demystifying}.
    \item Our approach transforms redundant per-thread global-memory accesses into efficient per-block loading, reducing LUT global-memory traffic by 64$\times$ per thread block.
    \item We provide comprehensive experimental evaluation across three models and diverse batch sizes, demonstrating 2.0--2.2$\times$ kernel speedup and up to 1.54$\times$ end-to-end improvement across model sizes from 27B to 70B parameters. The optimization reduces dequantization instruction count by 71\% through simplified indexing logic that eliminates warp divergence.
\end{itemize}

\section{Background}

\begin{figure*}
    \centering
    \includegraphics[width=0.8\textwidth, page=2]{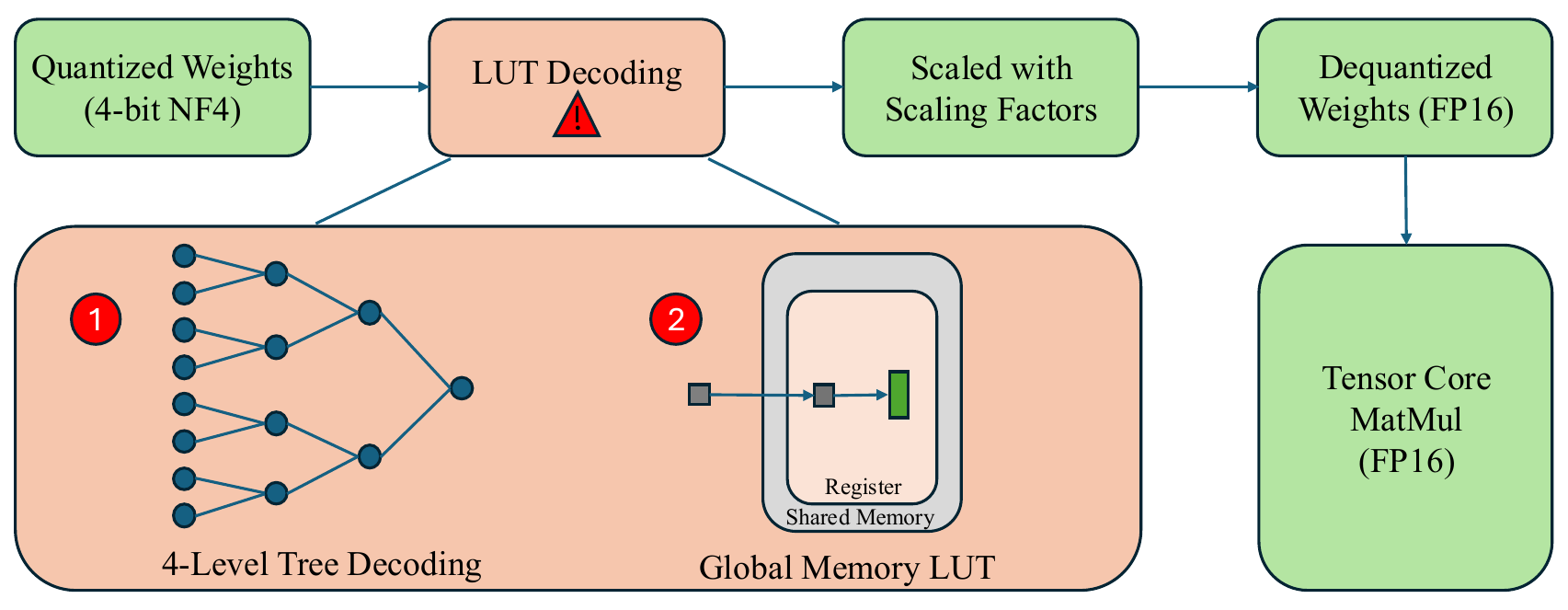}
    \caption{Baseline NF4 dequantization showing bottlenecks (1) 4-level tree decoding with branching overhead and warp divergence (2) Global memory code access at 290 clock cycles per operation}
    \label{fig:bottleneck}
\end{figure*}

\subsection{Large Language Model Inference}

Modern transformer-based language models consist of stacked decoder layers, with architectures ranging from 28 layers in smaller models to 64 layers in Qwen3-32B~\cite{qwen3_release}. Each layer contains self-attention and feed-forward network sublayers, involving matrix multiplications between input activations and learned weight matrices. For a model with $P$ parameters, the computational cost of a forward pass is approximately $2P$ FLOPs per token during autoregressive inference~\cite{kaplan2020scaling}. During autoregressive inference, the model processes tokens sequentially, with each layer's output serving as input to the next layer.

For an NF4 quantized model, the weight matrices in each layer require dequantization from 4-bit to FP16 format before computation on current GPU hardware. Each transformer layer contains multiple weight matrices: query, key, value, and output projections for attention, plus gate, up, and down projections for the feed-forward network. Processing input tokens through many layers (e.g., Qwen3-32B's 64 layers) requires dequantizing hundreds of weight matrices, with each matrix containing millions of parameters. Dequantization cost scales with model depth and the number of weight matrices per layer, but remains approximately constant per forward pass regardless of batch size, since only the fixed weight matrices---not the batch-dependent activations---require dequantization.

% \begin{figure}[b]
% \centering
% \includegraphics[width=1\textwidth, page=2]{Dequantization_Crop_Crop_Crop.pdf}
% \caption{Baseline NF4 dequantization showing bottlenecks
% % (1) 4-level tree decoding with branching overhead and warp divergence (2) Global memory LUT access at 290 clock cycles per operation
% }
% \label{fig:bottleneck}
% \end{figure}

\subsection{NF4 Quantization Fundamentals}

NF4 quantization addresses the memory challenge through a mathematically principled approach that aligns with the empirical distribution of neural network weights~\cite{dettmers2023qlora}. Unlike uniform quantization schemes that distribute levels evenly across the value range, NF4 concentrates precision around zero where most weights cluster. The scheme uses 16 carefully chosen quantization levels following quantile values of a theoretical normal distribution, normalized to fit the range [-1, 1]. NF4 uses a block size of 64 elements per scaling factor, balancing quantization granularity with metadata overhead. This information-theoretic approach minimizes quantization error for normally distributed weights while enabling efficient 4-bit storage.

Beyond inference optimization, NF4 has become essential for QLoRA (Quantized Low-Rank Adaptation), which enables fine-tuning of 65B+ parameter models on single consumer GPUs~\cite{dettmers2023qlora}. QLoRA combines NF4 quantization of base model weights with training low-rank adapter matrices in higher precision, reducing fine-tuning memory requirements from over 780GB to under 48GB for a 65B model. The base weights remain in 4-bit NF4 format during both forward and backward passes, with gradients computed only for the small adapter matrices. This makes NF4 not just an inference optimization but a foundational technology for democratized model customization~\cite{hu2021lora}.

\section{Motivation}

\subsection{The Quantization-Compute Gap}

The fundamental challenge stems from a critical architectural mismatch where memory capacity constraints demand 4-bit quantization, but compute units require 16-bit inputs. To run Qwen3-32B on an A100 GPU, the model requires 64GB in FP16 format but only 16GB when quantized to NF4, making quantization essential for single-GPU deployment. However, the A100's Tensor Cores only support FP16/BF16 operations, necessitating dequantization for every matrix multiplication.

This creates a bottleneck where dequantization overhead dominates inference time. Table~\ref{tab:baseline_overhead} shows baseline overhead percentage (OH\%) and kernel latency (KL) for Qwen3-32B across batch sizes on the GSM8K dataset.

\begin{table}[h]
\centering
\footnotesize
\begin{tabular}{|c|c|c|}
\hline
\multirow{2}{*}{\textbf{Batch Size}} & \multicolumn{2}{c|}{\textbf{GSM8K Dataset}} \\
\cline{2-3}
 & \textbf{OH\%} & \textbf{KL(ms)} \\
\hline
2 & 39.8 & 12983 \\
4 & 39.3 & 13031 \\
8 & 37.4 & 13021 \\
16 & 34.7 & 13046 \\
32 & 29.5 & 13041 \\
64 & 21.4 & 13027 \\
\hline
\end{tabular}
\caption{Baseline dequantization overhead (OH\%) and kernel latency (KL) for Qwen3-32B.}
\label{tab:baseline_overhead}
\end{table}

Dequantization overhead remains consistently high at 21-40\% of total inference time across batch sizes (2-64), with the baseline's reliance on global memory lookups resulting in approximately 13,000ms of consistent overhead. This validates shared memory optimization as our target.

\subsection{Memory Access Pattern Analysis}

The inefficiency of current dequantization implementations stems from suboptimal memory access patterns that fail to exploit GPU memory hierarchy. The baseline kernel loads the 16-element NF4 dequantization codes from global memory repeatedly for each thread's dequantization operations, resulting in redundant transfers of identical data. Global memory access on Ampere architecture~\cite{ampere_whitepaper} incurs 290 clock cycles, compared to 23 cycles for shared-memory reads and 19 cycles for writes~\cite{abdelkhalik2022demystifying}, creating a severe bottleneck for the billions of lookup operations required during inference.

The scattered access pattern and lack of data reuse at the thread block level mean the GPU's substantial compute capabilities remain underutilized while waiting for memory operations to complete. Our kernel-level profiling confirms this memory-bound behavior, with dequantization kernels consuming 21--40\% of end-to-end latency despite the relatively simple computational operations involved~\cite{pytorch_profiler}.

\subsection{Profiling Analysis and Bottleneck Identification}

For the Qwen3-32B batch-size-8 configuration in Table~\ref{tab:baseline_overhead}, PyTorch Profiler reveals the severity of the dequantization bottleneck in production workloads~\cite{pytorch_profiler}. Within quantized matrix multiplication, dequantization kernels account for 72.4\% of the time, with the remainder split between the underlying matmul computation and activation memory transfers. Note that this 72.4\% figure is scoped to quantized matmul (not end-to-end latency), and end-to-end latency also includes attention, KV-cache updates, sampling, and framework overhead, which is why Table~\ref{tab:baseline_overhead} reports 21--40\% dequantization overhead at the end-to-end level.

The baseline kernel's 4-level tree structure for NF4 dequantization requires multiple conditional branches and comparisons for each weight element, creating both computational overhead and warp divergence. Figure~\ref{fig:bottleneck} illustrates these bottlenecks: the tree-based decoding with branching overhead causes warp divergence as threads take different conditional paths, and global memory code access incurs 290 clock cycles per operation on Ampere architecture. This memory-bound behavior persists across all batch sizes, though the relative impact varies with workload characteristics.

\section{Problem Formulation and Design}

\subsection{Optimization Objectives and Constraints}

Our design targets three primary objectives while respecting critical constraints. First, we aim to minimize global memory accesses for lookup table operations by exploiting data reuse opportunities within thread blocks. Second, we seek to reduce instruction count through simplified indexing logic that eliminates complex branching and reduces bit manipulation to simple shifts and masks. Third, we must maintain token-exact compatibility with the baseline implementation to ensure correctness. These objectives must be achieved within the constraints of preserving ecosystem compatibility with BitsAndBytes and HuggingFace Transformers, requiring zero offline preprocessing or model conversion, and maintaining the modular kernel interface that enables composition with other optimizations.

\subsection{Shared Memory Loading Strategy}

The core insight of our optimization is that the 16-element NF4 code table, requiring only 64 bytes in FP32 format, fits comfortably in shared memory and can be reused across all threads in a block. We implement a single-thread loading strategy in which a single thread loads the lookup table into shared memory at kernel launch, followed by a synchronization barrier that ensures all threads see the complete table before proceeding with dequantization.

\begin{figure}[h]
\centering
\includegraphics[width=0.5\textwidth, page=3]{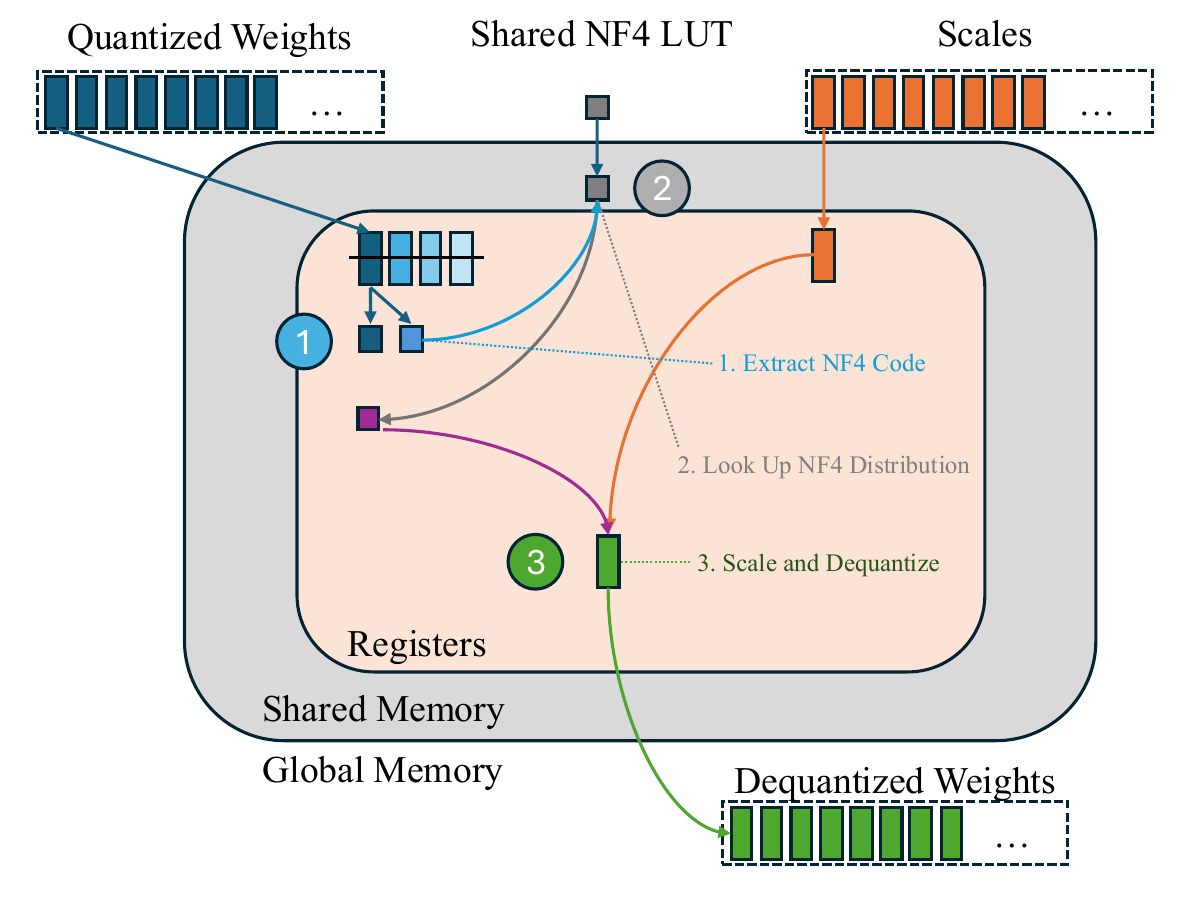}
\caption{Memory-level architecture with shared NF4 LUT serving all lookups within the thread block.}
\label{fig:memory_arch}
\end{figure}

This approach transforms the memory access pattern from one lookup-table load per thread (repeated across many dequantization operations) to a single load per thread block. For the kernel configuration with TILE\_SIZE=512, this represents a significant reduction in global memory traffic for lookup table access. With 64 threads per block processing 8 elements each (NUM\_PER\_TH=8), each block handles 512 weight elements in total. The shared memory's single-cycle broadcast capability allows multiple threads to read the same table entry simultaneously without bank conflicts, maximizing throughput. Figure~\ref{fig:memory_arch} illustrates this memory hierarchy optimization where the quantized weights and scales reside in global memory while the shared NF4 LUT in on-chip shared memory serves all lookups within the thread block, eliminating repeated global memory accesses that dominated the baseline implementation.

\subsection{Memory Loading vs. Hardcoded Literals}

We load the lookup table from constant memory into shared memory rather than hardcoding literals in the kernel. Hardcoding embeds float immediates into PTX instructions, increasing kernel code size and instruction cache pressure on the per-SM instruction cache (128 KB on Ampere~\cite{ampere_whitepaper}). Loading from global memory leverages the dedicated 64 KB constant cache per SM, which provides broadcast capability where a single memory read can serve all 32 threads in a warp simultaneously when they access the same address. This results in a more compact kernel binary with fewer instructions, improving instruction fetch efficiency and reducing stalls in high-occupancy scenarios. The constant memory declaration ensures robust performance across varying workload characteristics while minimizing register pressure.

\subsection{Simplified Index Computation}

Beyond memory optimization, we redesign the index computation logic to minimize instruction overhead. The baseline implementation employs a binary tree traversal for index calculation, requiring multiple conditional branches and arithmetic operations. Our approach leverages the regular structure of NF4 quantization to implement direct indexing through bit manipulation. This eliminates branch divergence within warps and reduces the indexing operation from 7 instructions to 2 instructions per weight. The simplified logic also improves instruction-level parallelism by removing data dependencies between successive index computations.

\subsection{Algorithm Design}

Algorithm~\ref{alg:dequant} presents the pseudocode for our optimized NF4 dequantization kernel. The key innovation lies in the single-thread loading of the NF4 lookup table into shared memory (lines 4-8), followed by direct indexed access during dequantization (lines 13-14) rather than tree-based branching used in the baseline implementation.

\begin{algorithm}[h]
\caption{Optimized NF4 Dequantization Kernel}\label{alg:dequant}
\begin{algorithmic}[1]
\State \textbf{Require:}
\State \hspace{0.5cm} $\bullet$ Quantized NF4 weights $A$, scaling factors $absmax$
\State \hspace{0.5cm} $\bullet$ Allocated shared memory $smem\_nf4[16]$
\If{$threadIdx.x = 0$}
    \For{$i = 0$ to $15$}
        \State $smem\_nf4[i] \gets nf4\_data[i]$ from constant mem
    \EndFor
\EndIf
\State $\_\_syncthreads()$
\For{$j = 0$ to $N_{\text{bytes}} - 1$ \textbf{in parallel}}
    \State $qval \gets A_{\text{nf4}}[j]$
    \State $scale \gets absmax[blockIdx]$
    \State $out[2j] \gets smem\_nf4[qval \gg 4] \times scale$
    \State $out[2j+1] \gets smem\_nf4[qval$ \& $0x0F] \times scale$
\EndFor
\State \textbf{Ensure:} Dequantized FP16 weights $out$
\end{algorithmic}
\end{algorithm}

The baseline implementation uses a 4-level binary tree with conditional branches, requiring up to 7 instructions per lookup with potential warp divergence. Our optimized version reduces this to 2 instructions: bit extraction and direct shared memory access. This transformation eliminates all branching in the critical dequantization loop, ensuring all threads in a warp execute identically.

\subsection{Thread-Level Architecture}

The thread-level implementation employs a single-thread loading strategy where thread 0 loads the 16-element NF4 lookup table (64 bytes) into shared memory while other threads wait at a synchronization barrier. This design was chosen over cooperative loading using many threads because the small data size makes the overhead of thread coordination and potential bank conflicts outweigh parallelism benefits. The 64-byte sequential copy latency is negligible compared to the thousands of subsequent shared memory accesses during dequantization.

The kernel is configured with 64 threads per block, processing 8 elements per thread (NUM\_PER\_TH=8), yielding a tile size of 512 elements. This configuration achieves good occupancy (2-3 blocks per SM on Ampere) without excessive register pressure. The 8 elements per thread provides sufficient work granularity to amortize instruction overhead while maintaining manageable register usage. The resulting 512-element tile size aligns well with typical weight matrix dimensions and cache line sizes.

During dequantization, all 64 threads simultaneously access the shared NF4 lookup table. Shared memory on Ampere supports broadcast reads—when multiple threads in a warp access the same address, the value is broadcast to all requesting threads in a single operation. This eliminates bank conflicts and provides 15$\times$ latency advantage (19 cycles for shared memory versus 290 cycles for global memory). The access pattern exhibits excellent temporal locality as threads repeatedly access the lookup table across 512 dequantization operations per thread block (64 threads $\times$ 8 elements), justifying the initial loading overhead.

\begin{figure}[h]
\centering
\includegraphics[width=0.5\textwidth, page=4]{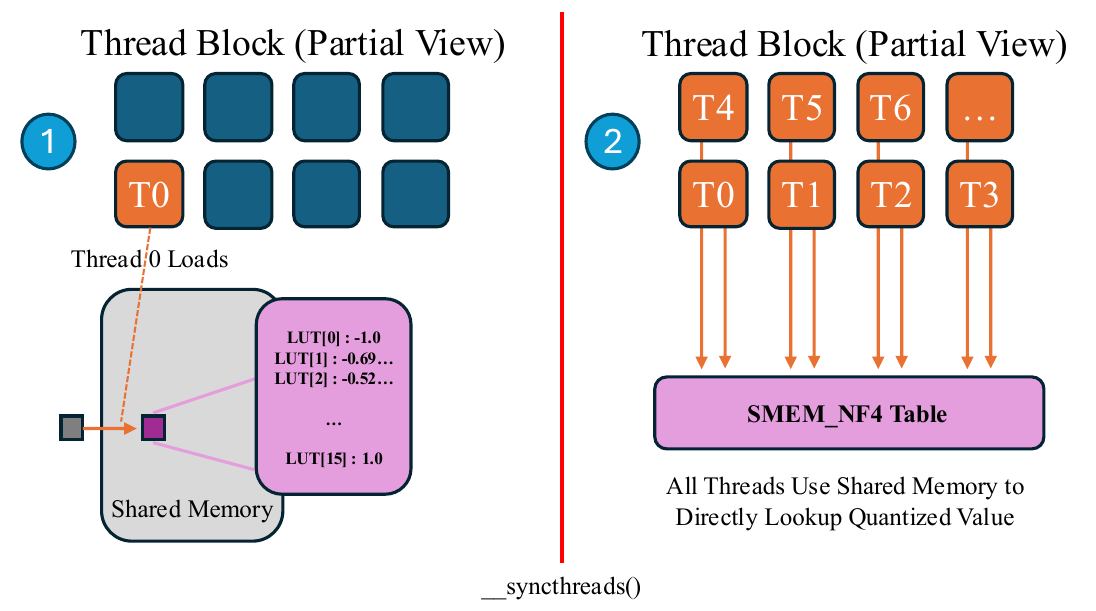}
\caption{Thread-level architecture with single-thread loading of LUT into shared memory.}
\label{fig:thread_arch}
\end{figure}

\section{Implementation}

\begin{figure*}[t]
\centering
    % Row 1: End-to-End Latency
    \begin{subfigure}[t]{0.32\textwidth}
        \centering
        \begin{tikzpicture}
            \begin{axis}[
                axis line style={line width=1pt},
                title={Gemma 27B},
                title style={yshift=-5pt},
                width=1.1\textwidth,
                height=4cm,
                ybar=0pt,
                bar width=.2cm,
                tick label style={font=\scriptsize},
                symbolic x coords={A, B, C, D, E, F},
                xticklabel style={yshift=5pt, align=center},
                xticklabels={2, 4, 8, 16, 32, 64},
                xtick=data,
                xtick style={draw=none},
                ytick={0, 5, 10, 15, 20, 25, 30},
                ymax=30,
                ymin=0,
                xlabel={Batch size},
                ylabel={E2E Latency (s)},
                ylabel style={yshift=-5pt},
                enlarge x limits=0.15,
                grid=major,
                grid style=solid,
                major x grid style={draw=none},
                legend entries={Baseline, Optimized},
                legend style={at={(0.5,0.98)}, anchor=north, font=\scriptsize, 
                              legend columns=2, column sep=3pt,
                              /tikz/every even column/.append style={column sep=3pt}},
                legend image code/.code={
                    \draw[#1] (0cm,-0.1cm) rectangle (0.3cm,0.1cm);
                },
            ]
            \addplot[fill=gray!40] coordinates {(A, 17.34) (B, 17.2) (C, 17.62) (D, 18.4) (E, 20.06) (F, 23.13)};
            \addplot[fill=teal!60] coordinates {(A, 17.08) (B, 17.16) (C, 17.06) (D, 17.04) (E, 17.15) (F, 17.49)};
            \end{axis}
        \end{tikzpicture}
    \end{subfigure}
    \begin{subfigure}[t]{0.32\textwidth}
        \centering
        \begin{tikzpicture}
            \begin{axis}[
                axis line style={line width=1pt},
                title={Qwen3 32B},
                title style={yshift=-5pt},
                width=1.1\textwidth,
                height=4cm,
                ybar=0pt,
                bar width=.2cm,
                tick label style={font=\scriptsize},
                symbolic x coords={A, B, C, D, E, F},
                xticklabel style={yshift=5pt, align=center},
                xticklabels={2, 4, 8, 16, 32, 64},
                xtick=data,
                xtick style={draw=none},
                ytick={0, 5, 10, 15, 20, 25, 30, 35},
                ymin=0,
                xlabel={Batch size},
                ylabel style={yshift=-12pt},
                enlarge x limits=0.15,
                grid=major,
                grid style=solid,
                major x grid style={draw=none},
            ]
            \addplot[fill=gray!40] coordinates {(A, 20.03) (B, 20.37) (C, 21.25) (D, 22.55) (E, 25.10) (F, 30.76)};
            \addplot[fill=teal!60] coordinates {(A, 18.69) (B, 18.40) (C, 18.59) (D, 18.82) (E, 19.43) (F, 24.05)};
            \end{axis}
        \end{tikzpicture}
    \end{subfigure}
    \begin{subfigure}[t]{0.32\textwidth}
        \centering
        \begin{tikzpicture}
            \begin{axis}[
                axis line style={line width=1pt},
                title={Llama3.3 70B},
                title style={yshift=-5pt},
                width=1.1\textwidth,
                height=4cm,
                ybar=0pt,
                bar width=.2cm,
                tick label style={font=\scriptsize},
                symbolic x coords={A, B, C, D, E, F},
                xticklabel style={yshift=5pt, align=center},
                xticklabels={2, 4, 8, 16, 32, 64},
                xtick=data,
                xtick style={draw=none},
                ytick={0, 10, 20, 30, 40, 50},
                ymin=0,
                xlabel={Batch size},
                ylabel style={yshift=-12pt},
                enlarge x limits=0.15,
                grid=major,
                grid style=solid,
                major x grid style={draw=none},
            ]
            \addplot[fill=gray!40] coordinates {(A, 39.96) (B, 40.55) (C, 41.64) (D, 41.89) (E, 43.75) (F, 46.3)};
            \addplot[fill=teal!60] coordinates {(A, 26) (B, 26) (C, 26.38) (D, 27.1) (E, 29.35) (F, 32.18)};
            \end{axis}
        \end{tikzpicture}
    \end{subfigure}
    
    % Row 2: Throughput
    \begin{subfigure}[t]{0.32\textwidth}
        \centering
        \begin{tikzpicture}
            \begin{axis}[
                axis line style={line width=1pt},
                width=1.1\textwidth,
                height=4cm,
                ybar=0pt,
                bar width=.2cm,
                tick label style={font=\scriptsize},
                symbolic x coords={A, B, C, D, E, F},
                xticklabel style={yshift=5pt, align=center},
                xticklabels={2, 4, 8, 16, 32, 64},
                xtick=data,
                xtick style={draw=none},
                ytick={0, 100, 200, 300, 400, 500, 600, 700},
                ymin=0,
                xlabel={Batch size},
                ylabel={Throughput (tokens/s)},
                ylabel style={yshift=-5pt},
                enlarge x limits=0.15,
                grid=major,
                grid style=solid,
                major x grid style={draw=none},
            ]
            \addplot[fill=gray!40] coordinates {(A, 17.3) (B, 37.9) (C, 79.9) (D, 153) (E, 284) (F, 506.3)};
            \addplot[fill=teal!60] coordinates {(A, 17.9) (B, 38) (C, 79.3) (D, 161.2) (E, 323.3) (F, 632.6)};
            \end{axis}
        \end{tikzpicture}
    \end{subfigure}
    \begin{subfigure}[t]{0.32\textwidth}
        \centering
        \begin{tikzpicture}
            \begin{axis}[
                axis line style={line width=1pt},
                width=1.1\textwidth,
                height=4cm,
                ybar=0pt,
                bar width=.2cm,
                tick label style={font=\scriptsize},
                symbolic x coords={A, B, C, D, E, F},
                xticklabel style={yshift=5pt, align=center},
                xticklabels={2, 4, 8, 16, 32, 64},
                xtick=data,
                xtick style={draw=none},
                ytick={0, 100, 200, 300, 400, 500, 600, 700},
                ymin=0,
                xlabel={Batch size},
                ylabel style={yshift=-12pt},
                enlarge x limits=0.15,
                grid=major,
                grid style=solid,
                major x grid style={draw=none},
            ]
            \addplot[fill=gray!40] coordinates {(A, 17.57) (B, 34.57) (C, 83.60) (D, 167.57) (E, 283.10) (F, 497.30)};
            \addplot[fill=teal!60] coordinates {(A, 18.97) (B, 38.57) (C, 96.13) (D, 189.93) (E, 367.80) (F, 639.43)};
            \end{axis}
        \end{tikzpicture}
    \end{subfigure}
    \begin{subfigure}[t]{0.32\textwidth}
        \centering
        \begin{tikzpicture}
            \begin{axis}[
                axis line style={line width=1pt},
                width=1.1\textwidth,
                height=4cm,
                ybar=0pt,
                bar width=.2cm,
                tick label style={font=\scriptsize},
                symbolic x coords={A, B, C, D, E, F},
                xticklabel style={yshift=5pt, align=center},
                xticklabels={2, 4, 8, 16, 32, 64},
                xtick=data,
                xtick style={draw=none},
                ytick={0, 50, 100, 150, 200, 250, 300, 350},
                ymin=0,
                xlabel={Batch size},
                ylabel style={yshift=-12pt},
                enlarge x limits=0.15,
                grid=major,
                grid style=solid,
                major x grid style={draw=none},
            ]
            \addplot[fill=gray!40] coordinates {(A, 7.4) (B, 16.1) (C, 33.3) (D, 64.5) (E, 120.6) (F, 222.1)};
            \addplot[fill=teal!60] coordinates {(A, 11.4) (B, 24.8) (C, 50.5) (D, 96.4) (E, 175) (F, 305.2)};
            \end{axis}
        \end{tikzpicture}
    \end{subfigure}

    \caption{End-to-end latency (top row) and throughput (bottom row) comparison across three models showing consistent improvements with our optimized NF4 dequantization kernel.}
    \label{fig:performance_comparison}
\end{figure*}
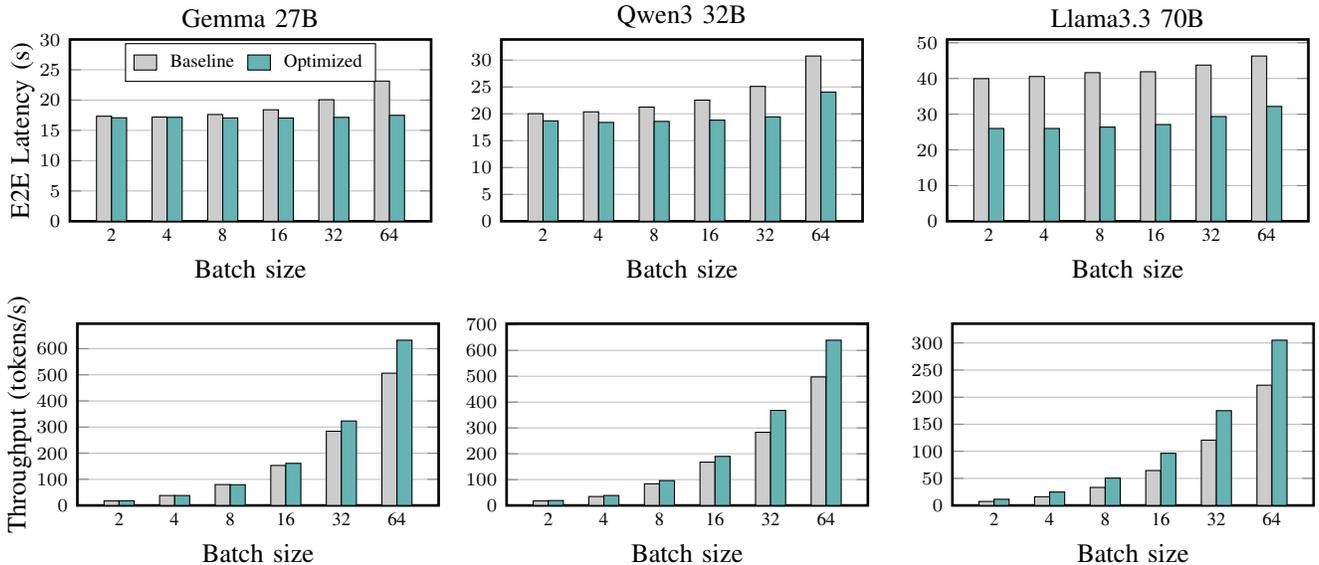

\subsection{Kernel Architecture}

Our optimized kernel modifies the \texttt{kDequantizeBlockwise} function in BitsAndBytes, instantiated with template parameters for FP16 output, 512-element tile size, 64 threads per thread block, 8 elements per thread, and NF4 data type. The implementation maintains full API compatibility with HuggingFace Transformers.

The kernel declares the NF4 lookup table in constant memory, enabling hardware-accelerated broadcast reads from the 64 KB constant cache per SM. Thread 0 copies the table to shared memory, followed by \texttt{\_\_syncthreads()} for visibility. The dequantization loop extracts 4-bit indices using bitwise operations (shift right by 4 for high nibble, AND with 0x0F for low nibble) and performs shared memory lookups. The baseline implementation uses a 4-level conditional tree requiring multiple branch instructions per weight; our bit manipulation compiles to 2 instructions with no branches, eliminating warp divergence.

Global memory accesses use CUB library primitives for coalesced transfers. Edge cases are handled via predicated execution when weight matrix size is not divisible by 512. The modified kernel integrates directly into BitsAndBytes and is accessed through the standard quantized linear layer API.

\subsection{Instruction-Level Optimizations}

Our optimization eliminates the baseline kernel's branching tree traversal for index computation. The baseline implementation requires multiple conditional branches to traverse a 4-level decision tree for each weight element, causing warp divergence when threads process different indices. Our approach uses direct bit manipulation with AND and shift operations to extract the 4-bit index, reducing the indexing operation from 7 instructions with branches to 2 instructions without branches.

This elimination of branch divergence is particularly important as all threads within a warp now execute the same direct indexing instruction rather than potentially taking different branches in the tree traversal. This maintains the GPU's SIMT execution model efficiently and prevents the serialization that occurs when threads within a warp follow different control flow paths.

\section{Experiments}

\subsection{Experimental Setup}

We conducted comprehensive testing on the USC CARC system using a single NVIDIA A100-80GB GPU paired with AMD EPYC 7513 CPU (32 cores) and 64GB RAM. The software environment consists of CUDA 12.6, PyTorch 2.1, and BitsAndBytes 0.47.0 with fixed random seeds ensuring reproducibility~\cite{cuda_programming_guide}. We evaluate our optimization on three models: Gemma 27B, Qwen3 32B, and Llama3.3 70B, with token-by-token accuracy verification confirming bit-exact outputs between baseline and optimized implementations for all models.

Our measurement protocol implements controls for accurate and reproducible results. Each experiment begins with a warmup phase of one inference pass to stabilize GPU clocks and populate caches. For optimized-kernel analysis, we run three profiling-enabled passes, reporting mean latency. We lock the GPU to its maximum clock (1410 MHz) using nvidia-smi, eliminating frequency scaling variance. The PyTorch Profiler provides microsecond-resolution timing via NVIDIA's CUPTI interface, including kernel launch overhead and memory transfers to reflect realistic end-to-end performance~\cite{pytorch_profiler}.

We evaluate performance across batch sizes 2, 4, 8, 16, 32, and 64 using input prompts from the GSM8K dataset~\cite{gsm8k} for all three models, comparing baseline dequantization against our shared memory optimized version. Each batch size is tested with multiple prompt lengths to ensure robustness across different workload characteristics.

\subsection{End-to-End Performance Results}

Our evaluation demonstrates consistent performance improvements across all three models and tested batch sizes (Figure~\ref{fig:performance_comparison}). The end-to-end speedup varies significantly by model size, with larger models demonstrating greater benefits: Llama3.3 70B achieves 1.52$\times$ average speedup across all batch sizes, Qwen3 32B achieves 1.18$\times$ average, and Gemma 27B achieves 1.10$\times$ average. This pattern occurs because larger models spend a higher proportion of total inference time in dequantization operations, increasing the impact of our 2.0-2.2$\times$ kernel-level improvements.

Figure~\ref{fig:performance_comparison} (top row) shows end-to-end latency across the three models. Llama3.3 70B demonstrates the most dramatic improvements, achieving 1.54$\times$ speedup at batch 2 (39.96s to 26s) and maintaining 1.44$\times$ speedup even at batch 64. Qwen3 32B shows strong mid-range performance with peak gains of 1.29$\times$ at batch 32, while Gemma 27B exhibits increasing benefits at larger batch sizes, reaching 1.32$\times$ at batch 64 (23.13s to 17.49s).

The varying end-to-end impact across batch sizes reflects how dequantization's proportion of total inference time changes with workload characteristics. Qwen3 32B profiling (Table~\ref{tab:baseline_overhead}) shows dequantization consumes 30-40\% of total inference time across batch sizes. At smaller batch sizes, memory bandwidth underutilization for matrix operations allows dequantization to dominate, enabling our kernel optimization to deliver greater end-to-end improvements. At larger batch sizes, improved matrix multiplication parallelization reduces dequantization's relative contribution, though it remains a significant bottleneck. Our consistent 2.0-2.2$\times$ kernel speedup thus translates to end-to-end improvements ranging from 1.07$\times$ to 1.54$\times$ depending on model architecture and batch configuration.

Figure~\ref{fig:performance_comparison} (bottom row) shows throughput improvements in tokens per second. The optimized implementation delivers consistent gains across all models and batch sizes. Llama3.3 70B shows the strongest throughput improvements with 1.54$\times$ at batch 2, leveraging its greater dequantization overhead to achieve substantial acceleration. Qwen3 32B achieves 1.30$\times$ improvement at batch 32 (283 to 368 tokens/s), while Gemma 27B demonstrates increasing gains with batch size, reaching 1.25$\times$ at batch 64 (506 to 633 tokens/s). These throughput improvements directly translate to production benefits, enabling higher system capacity and improved cost-effectiveness for deployed LLM inference systems.

This analysis demonstrates that dequantization remains a significant bottleneck across all practical batch sizes and model scales, validating our optimization target. The consistent kernel-level speedup translates to varying end-to-end improvements based on model architecture and workload characteristics, with larger models showing greater benefits due to their higher proportion of time spent in dequantization operations.

\subsection{Kernel-Level Performance Analysis}

The dequantization kernel demonstrates improvements across all configurations. Table~\ref{tab:kernel_results} presents the kernel-level speedup across all three models and batch sizes for the NF4 dequantization kernel. The optimization achieves consistent 2.0--2.2$\times$ speedup across all models. This consistency across different model sizes validates that our optimization targets fundamental memory access bottlenecks rather than model-specific characteristics.

\begin{table}[h]
\centering
\small
\begin{tabular}{|c|c|c|c|}
\hline
\textbf{Batch} & \textbf{Gemma} & \textbf{Qwen3} & \textbf{Llama3.3} \\
\textbf{Size} & \textbf{27B} & \textbf{32B} & \textbf{70B} \\
\hline
2 & 2.10$\times$ & 2.20$\times$ & 2.04$\times$ \\
4 & 2.10$\times$ & 2.19$\times$ & 2.04$\times$ \\
8 & 2.11$\times$ & 2.19$\times$ & 2.04$\times$ \\
16 & 2.10$\times$ & 2.19$\times$ & 2.03$\times$ \\
32 & 2.11$\times$ & 2.19$\times$ & 2.05$\times$ \\
64 & 2.08$\times$ & 2.15$\times$ & 2.03$\times$ \\
\hline
\textbf{Average} & \textbf{2.10$\times$} & \textbf{2.19$\times$} & \textbf{2.04$\times$} \\
\hline
\end{tabular}
\caption{Kernel-level dequantization speedup across models and batch sizes, demonstrating consistent 2.0--2.2$\times$ improvements.}
\label{tab:kernel_results}
\end{table}

The consistency of the speedup across both models and batch sizes confirms that our optimization targets fundamental memory access bottlenecks rather than model-specific or batch-specific effects. The slight variations between models (2.04--2.19$\times$) reflect differences in overall memory access patterns and compute intensity, but all three models benefit substantially from the shared memory optimization. The simplified indexing logic reduces the index computation from 7 instructions with branches to 2 instructions without branches, while shared memory access eliminates repeated global memory loads for the lookup table. Warp execution efficiency improves significantly due to the elimination of branch divergence, as all threads now execute the same direct indexing instruction rather than potentially taking different branches in the tree traversal.

The consistent improvements across different batch sizes indicate that the optimization scales effectively with varying workload characteristics, making it suitable for diverse deployment scenarios from research experimentation to production inference systems.

\section{Conclusion}

We present an optimized NF4 dequantization kernel achieving 2.0--2.2$\times$ kernel speedup and up to 1.54$\times$ end-to-end improvement through principled GPU memory hierarchy exploitation. By loading the 16-element NF4 lookup table into shared memory and simplifying index computation through direct bit manipulation, our lightweight approach requires only 64 bytes of shared memory per thread block while delivering consistent performance gains across models ranging from 27B to 70B parameters. The optimization maintains full compatibility with the HuggingFace ecosystem and requires no offline pre-processing or model conversion, enabling immediate deployment in production systems.

\vspace{12pt}

\end{document}